\documentclass[a4paper]{jpconf} 
\usepackage{graphicx}
\usepackage[dvipsnames]{xcolor}
\usepackage{amsmath}

\begin{document}
\title{Physics-Enhanced Machine Learning: a position paper for dynamical systems investigations}

\author{Alice Cicirello}

\address{Department of Engineering, University of Cambridge, Trumpington Street, CB2 1PZ, Cambridge}

\ead{ac685@cam.ac.uk}

\begin{abstract}
This position paper takes a broad look at Physics-Enhanced Machine Learning (PEML)  - also known as Scientific Machine Learning -  with particular focus to those PEML strategies developed to tackle dynamical systems' challenges.  The need to go beyond Machine Learning (ML) strategies is driven by: (i) limited volume of informative data, (ii) avoiding accurate-but-wrong predictions; (iii) dealing with uncertainties;  (iv) providing Explainable and Interpretable inferences. A general definition of PEML is provided by considering four physics and domain knowledge biases, and three broad groups of PEML approaches are discussed: physics-guided, physics-encoded and physics-informed. The advantages and challenges in developing PEML strategies for guiding high-consequence decision making in engineering applications involving complex dynamical systems, are presented. 
\end{abstract}

\section{Introduction}
Dynamical system models are used in  engineering and applied science fields to identify, analyze, control and predict a wide range of behaviours of real-world systems which are typically composed by many components, which may interact with each other, and for which a linear time-invariant modelling assumption is often inadequate \cite{Schouken}. Systems are objects producing a signal (output), depending on other signals (inputs and disturbances) and depending on some initial conditions. Dynamical systems can be used to investigate problems in applied mechanics and structural dynamics that are usually multi-scale, multi-physics, high-dimensional, strongly nonlinear, and time-dependent \cite{Schouken, keith, ref2}. 
Machine Learning (ML) approaches provide a very powerful set of tools for building the nonlinear dynamical systems model directly from observations. However, in many engineering applications, data is typically sparse, noisy, expensive to acquire,  and ultimately quite limited. The resulting ML models based on data-only cannot be fully deployed for guiding high-consequence decision making because they lack (i) generalisation to previously unseen conditions; (ii) interpretability; and (iii) robustness \cite{SciMLfirst}. 

Scientific Machine Learning (SciML) was first mentioned in the report summarising the results of a workshop \cite{SciMLfirst}. In general, SciML refers to the combination of computational science and ML approaches to leverage existing knowledge and physics models within learning schemes \cite{SciMLlastwork}. Much of the efforts in SciML have focused on accelerating solvers and on constraining ML-predictions by incorporating physics biases on data-driven architectures \cite{ref2,SciMLlastwork, paperPINNs}, consequently, Physics-Informed Neural Networks (PINNs) architectures \cite{ref2,paperPINNs, firstkarn} are sometimes wrongly perceived as the only SciML learning scheme. The term Physics-Enhanced Artificial Intelligence (PEAI) was introduced in \cite{PEAI} for ``intelligently combining models from the domains of artificial intelligence or machine learning with physical and expert models". The distinction between Physics-Guided Neural Networks (NNs), Physics-Informed NNs and Physics-Encoded NNs was presented in \cite{ref5}. Informed Machine Learning (IML) was introduced in \cite{InfML} to describe prior knowledge integration into the learning process, and a taxonomy was developed to enable a structured categorization of different approaches.  Since 2019, the term Physics-Enhanced Machine Learning (PEML) has been used in a broad sense to encompass strategies where ``prior physics knowledge is embedded to the learner" \cite{eleni}, especially for dealing with dynamical systems in Engineering. 

It is worth noting that development of hybrid physics-data models to tackle real-world Engineering problems  have been explored for more than two decades. For example,  in Structural Health Monitoring (SHM) \cite{keith, ref4, ref4bis, eleni} and in Nonlinear System Identification of dynamical systems \cite{Schouken}. One of the fundamental building blocks for these hybrid physics-data approaches has been the paper on ``Bayesian calibration of computer models" \cite{KH},  where Bayesian Inference is combined with physics-based models (likelihood) and domain knowledge (prior distribution). Recent advances in ML have lead to innovative ways to build hybrid physics-data models for addressing challenges in dynamical systems.

A general definition of Physics-Enhanced Machine Learning is provided in what follows, encompassing previous definitions of PEML, SciML, IML and PEAI. This is based on considering four physics and domain knowledge biases, and it enables a framework where it is possible to distinguish three broad groups of hybrid physics-data model strategies: Physics-Guided, Physics-Informed and Physics-Encoded approaches. Similar or alternative groupings have been considered, as discussed for example in \cite{ref4, ref4bis, eleni}.

\section{Why is there a need to go beyond data-driven for dynamical systems?}
ML strategies initially developed for computer science applications \cite{bishop,DLbengio}, have seen an explosive growth in their use across various fields because of (i) the use of GPUs to accelerate their execution, (ii) the availability of ``big data", and (iii) the many ML libraries freely available. 
While in traditional Engineering modelling the machine (i.e., the computer) is used to obtain the output data from a given Algorithm (a physical relationship between causes and effects, i.e., a physics-based model) and input data, both input and output data are used in ML so that machines can discover representative algorithms, and as a result these approaches are often referred to as data-driven. This is schematically shown in Figure 1. 

\begin{figure}[htb!]
\centering
\includegraphics[width=26pc]{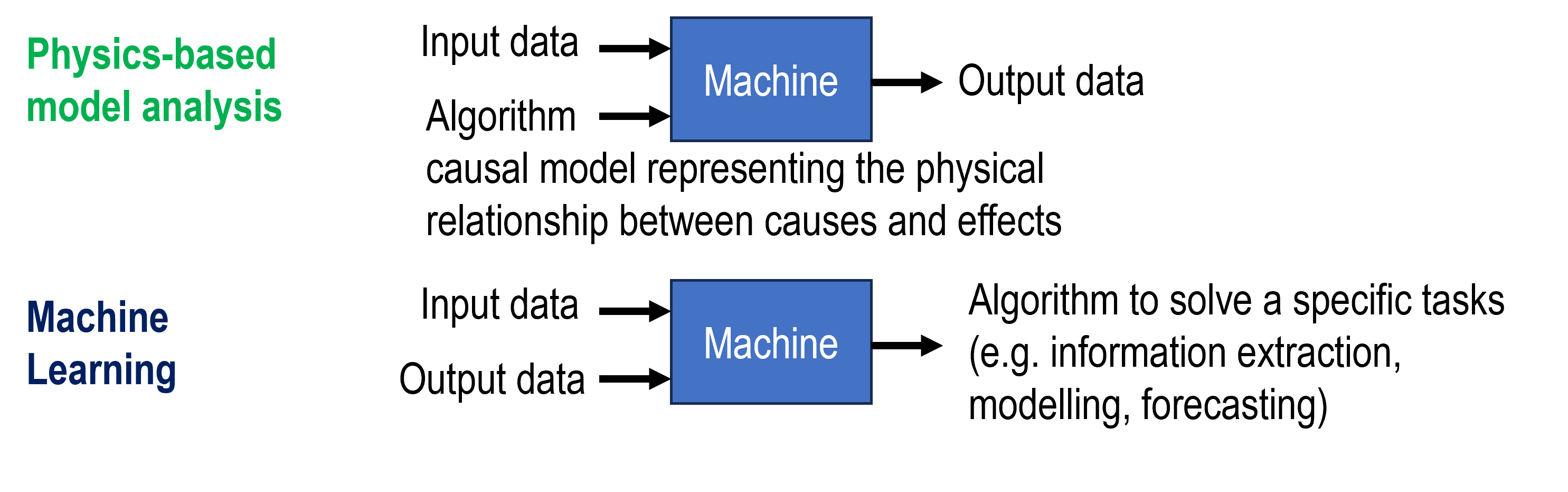}
\caption{Physics-based model analysis versus Machine Learning.}
\end{figure}
SHM \cite{keith}, together with the field of Nonlinear System Identification \cite{Schouken}, have been among the engineering fields that have seen an early adoption of ML strategies to make inferences on the (health) status of structures in operating conditions and to guide decision making on structural dynamical systems.  Nowadays, ML strategies are widely used in engineering for a broad range of tasks, including but not limited to: (i) developing models from data; (ii) reducing computational costs; (iii) extracting and summarising information from large datasets; (iv) learning patterns, trends and correlations; (v) inspiring the developments of new experiments; (vi) online tracking and learning system's behaviour; (vii) guiding decision making.

ML algorithms produce specialised models investigating features (patterns in the data), correlations, relationships and dependencies, associational evidence and hidden structures. However, typically no causal relationships between variables are accounted for (unless causal inference approaches combining machine learning and causal inference methods are employed \cite{causal}) and confounding sources are not automatically disentangled (unless advanced approaches for representation learning are implemented \cite{disent}). Moreover, these data-driven models will be as good as the data provided to train them. The applications of ML techniques to Engineering for guiding high consequence decision making,  and especially those involving dynamical systems, poses several challenges. The most critical ones are discussed in what follows.

\subsection{Challenge one: the volume of useful data is generally limited}

 Some assumptions made on the available data used to build data-driven models, which are often forgotten, are: (i) data is \textit{informative}, that is the data can be used for a specific task (e.g., classification, regression, and so on) since important patterns (features) characterising a phenomenon can be extracted from the raw data. These features would be informative, discriminating, and independent (e.g., 
environmental and operational conditions can be distinguished from health conditions of the structure and/or of the monitoring system given a set of raw data). In engineering applications this poses the challenges of what data should be used, that corresponds to what to measure, when and where. (ii) data is \textit{accurate}, that is data/measurements are close to their true value (e.g., low noise levels, sensors are calibrated) and free from data-biases (data is collected in a clear and consistent manner); (iii) data is \textit{reliable}, that is if under the same circumstances, the same behaviour will be observed consistently over an extended period. This also means that the data must be up-to-date.

These assumptions are often not met by the large volumes of data collected in Engineering applications, especially when dealing with dynamical systems that interact with time-varying environmental conditions, changing operating conditions and subject to degradation and deterioration phenomena and/or maintenance activities which can modify the underlying dynamical system behaviour (which have the potential of making part of the data collected for the same structure obsolete depending on the task at hand). Moreover, informative datasets are expensive to acquire, and they often come from heterogeneous sources with different temporal and spatial resolutions and with different accuracy. In addition to these, not all the conditions that could potentially arise in a structure can be observed and/or labelled (e.g., damage conditions). Furthermore, failures of the monitoring system might lead to data-biases (e.g., noisy, sparse, gappy, inaccurate data). 

Data cleaning is one of the most important and crucial steps when handling data \cite{dataclean}, especially for engineering applications. Data needs to be standardised; duplicates and redundant observations must be removed. To avoid distorting the distribution of the data, missing data handling must be carried out. Structural errors related to merging datasets collected at different times from different databases (e.g., same attribute with a partially different name) often have to be handled manually. Finally, outliers' management must be carried out by removing unwanted values not fitting the dataset. Consequently, even if terabytes of data are collected for investigating complex dynamical systems, the volume of useful data is usually limited. Therefore, severely limiting the range of applicability of data-driven only approaches for some dynamical systems especially in structural dynamics applications.

\subsection{Challenge two: avoiding accurate-but-wrong predictions}
One of the biggest challenges in engineering applications is avoiding models that lead to accurate-but-wrong predictions.  The well-known quote by George E.P. Box states: ``All models are wrong, but some are useful''. It is important to choose a modelling strategy based on the available information, and being aware about the assumptions made. Since the model represents only some “aspects” of the real world, it should be used within its range of validity. This often means understanding with respect to what dataset the model has  been validated.  

The blind application of data-driven models to support decision making in engineering can be ``dangerous''. When considering complex dynamical systems, it is often not possible to fully measure and characterise the Environmental and Operational loading acting on these systems, or obtaining output measurements with a detailed spatial and/or temporal resolution. The data-driven models developed would be therefore based on correlations and associational evidence, and not specifically accounting for causal relationships or distinguishing confounding sources. Consequently, data-driven models might lead to: (i) poor generalization performance to previously unseen (and/or unknown) conditions - for example,  because they might be used outside the validated dataset (i.e., extrapolation) and/or because they were overparametrized (i.e., overfitting); and (ii)  physically inconsistent or implausible predictions - for example, because of observational biases caused by corrupted or not-informative measurements (i.e wrong interpolation). Moreover, these models might lead to accurate-but-wrong predictions because of uncertainties in the observations, in the architecture setup, and in the model hyperparameters. 

It is worth mentioning that also physics-based models can be ``dangerous''. The fidelity and complexity of the model depends on the task. Advanced physics-based model might have strong generalization performance at the price of expensive and time-consuming modelling setup and high computational cost for generating outputs. These models would lead to physically consistent but not necessarily plausible predictions, depending on the simplifying assumptions made on the model and model parameters. Moreover, in structural dynamics applications, these models might lead to accurate-but-wrong predictions even if the physics of the problem is correctly specified because of uncertainty in the loading, in the material properties, geometry, boundary conditions, constitutive laws and in the nonlinearity parameters accounted for.   Indeed, the combination of physics-based model and domain knowledge with data-driven techniques could improve the predicting capabilities of both models used in isolation, but it would still require to characterise the different sources of uncertainties.

\subsection{Challenge three: dealing with uncertainty}
Let us consider a complex  dynamical system representing a wind turbine in operating conditions subject to some Environmental and Operational (E\&O) loading (inputs) which will affect its true dynamic response over time ($y_{true}$). The complete input $(\bf{x})$  - output ($y_{true}$) relationship $f(\bf{x})$ is usually not fully known because of uncertainties in the system (e.g., material and geometric properties, boundary conditions), inevitable manufacturing variability and differences between the properties of the designed and as-built structure. As a result, the manufactured/assembled/built system's behaviour often deviates significantly from the virtual prototype (template digital twin) and/or from the physical prototype usually tested in controlled laboratory conditions.

\begin{figure}[htb!]
\centering
\includegraphics[width=33pc]{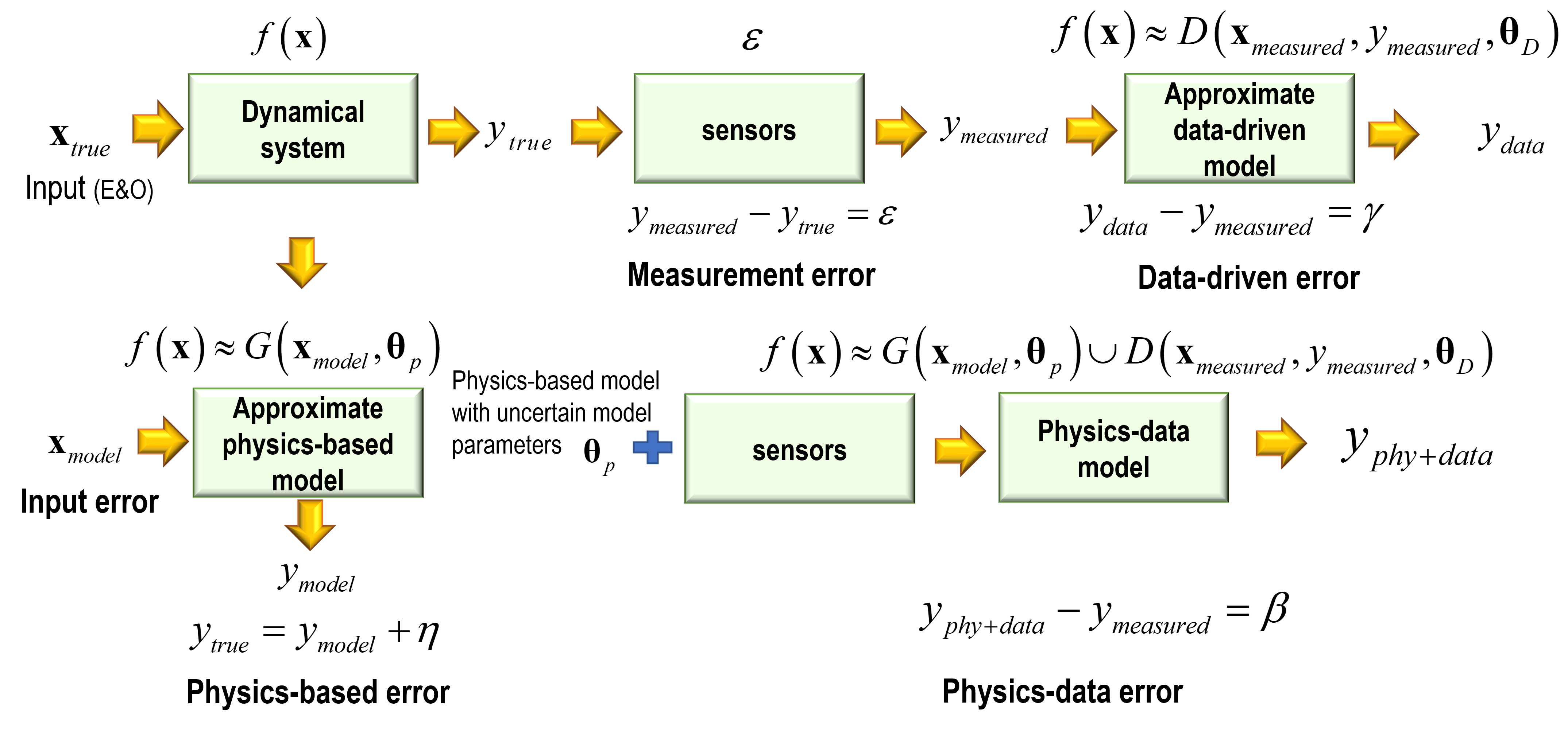}
\caption{Errors introduced when building models of a dynamical system}
\end{figure}

As schematically shown in Figure 2, sensors can be used to obtain measurements of the wind turbine. However, each $y_{measured}$ would be different from $y_{true}$ since observations are affected measurements errors caused for example by instrument measurement noise and/or systematic errors or biases (e.g., calibration issues, sensor drift). A ML algorithm can then be used to build, for example, a regression model of the measurements or to build an approximation of the $f(\bf{x})$ using a set of recorded inputs and outputs. However, the full characterization of the inputs (and outputs) is usually not possible: they are time-varying, non-stationary, affected by noise, difficult to be directly measured and if reconstructed, they might be obtained with an insufficient spatial and temporal resolution, some conditions might not be observed (incomplete domain coverage) and it might not be possible to disentangle the contributions of different inputs. Depending on the data-driven techniques selected (e.g., Artificial Neural Networks, Gaussian Process (GP) \cite{bishop}), informative data availability and hyperparameters ($\boldsymbol{ \theta}_D$)  of the data-driven model identified, there would be a data-driven error. This data-driven error can be split into two parts:  ``model structural errors" \cite{Schouken} and model parameters. To avoid confusion with the physics-based model of the structure,  ``model structural errors" are going to be referred to as model form errors. Model form errors account for the many plausible model choices and model parametrization, caused by limited and incomplete noisy data, as well as errors in the computer implementation/solver. The model parameters include those estimated from data automatically and model hyperparameters (set manually and to be tuned to estimate model parameters). 

Alternatively, one might use a physics-based model of the dynamical system (for example, the one used at the design stage) and consider that some of the parameters of this model might be uncertain (e.g., material properties, geometry, boundary conditions, constitutive laws and nonlinearity). The results obtained with this model would be different from the actual dynamical system, because of a physics-based error. Physics-based error could be caused by incorrect assumptions on the system behaviour, e.g., Linear Time Invariant (LTI) system which disregards the effects of nonlinearity and of time-varying phenomena such as deterioration and/or degradation. Similarly to the data-driven error, the physics-based error could be split into two parts, one related to the model form (caused by using models not fully capturing the physics of the problem at hand and/or errors in the computer implementation/solver)  and the other one related to the parameters of the physics-based model.  Moreover, the input might be modelled with some simplifying assumptions leading to an input error.

It is important to stress that using a data-driven or a physics-based model would change the way we deal and account for uncertainty.  If informative data is available together with a detailed physical model, one could seek to build a hybrid physics-data model with the goal of updating some of the latent  parameters of a physics-based model. These approaches stem from Probabilistic Machine Learning strategies \cite{probML}, and in engineering applications they are usually referred to as probabilistic model updating strategies \cite{KH, lye}. Many physics-data models have been used for many years for tackling challenges in SHM \cite{keith, ref4, ref4bis, eleni} and in Nonlinear System Identification of dynamical system \cite{Schouken} to account for uncertainties. It is worth noting that using a hybrid physics-data model will lead to a combined physics-data error which can be also split into two parts: one related to the hybrid physics-data model form (e.g., likelihood function selected to describe the discrepancy between the observed informative data and model prediction \cite{KH}) and the other one related to the parameters of such model.  Alternative ways could be explored to combine the known physics and domain knowledge with the data, starting from a data-driven model or combining directly physics-models with data-driven models. Each model structure might lead to a different way of accounting for and quantifying uncertainty. 
The interested reader is referred to  \cite{Zou}  to learn more about the characterization of several source of errors when building a joined physics-data model of an offshore wind turbine. 

We can further analyse the different errors, as schematically summarised in Figure 3. Errors in physical parameters values (of a physics or hybrid physics-data model) can be caused by an intrinsic variability, usually referred to as aleatory uncertainty, or because of lack of knowledge, usually referred to as epistemic uncertainty \cite{KIUREGHIAN2009105,Uncertainty_dimensions}. As a result, if sufficient informative data would be available, one could reduce epistemic uncertainty on a parameter to a single value. Aleatory uncertainty instead is often referred to as irreducible, because even if sufficient informative data would be available, this would result in a more accurate distribution of the uncertain parameters, but not a single value.

\begin{figure}[htb!]
\centering
\includegraphics[width=33pc]{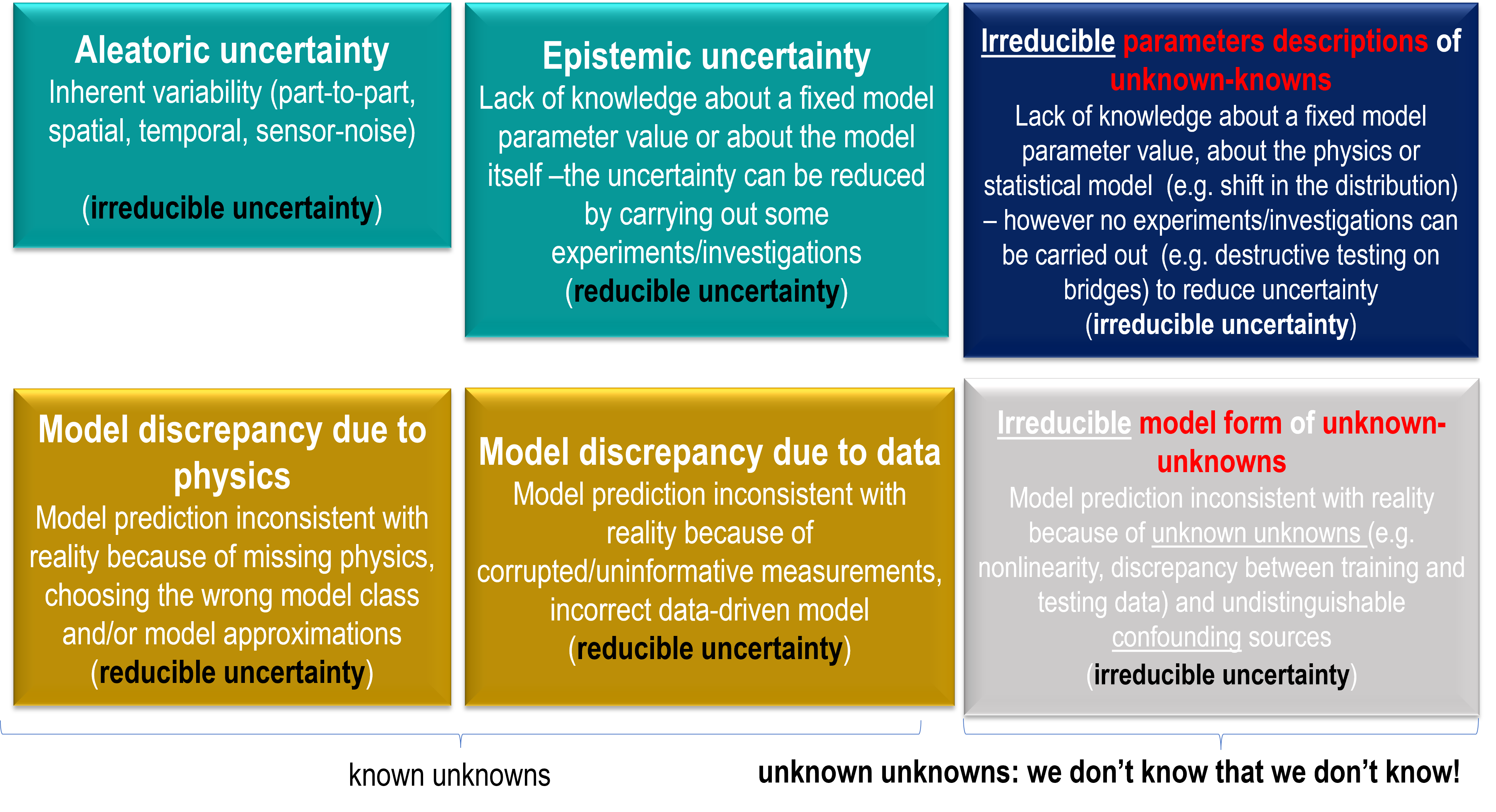}
\caption{Reducible and irreducible uncertainties}
\end{figure}

When focusing on the model errors, the uncertainty is referred to as model discrepancy. Depending on the use of a data-driven or physics based or hybrid physics-data models, there will be different model discrepancies.  The model discrepancies are reducible, in principle.  However, there might be instances where the model parameters and the model form uncertainties are irreducible. These irreducible uncertainties are often referred to as unknown-unknowns, and they can be broadly caused by (i) not being able to reduce/appropriately describe uncertainty (because of cost, time, impossibility of destructive testing or monitoring, access to experts and/or information, unavailability of informative data, and so on) or (ii) not accounting for or distinguishing  confounding sources of uncertainty because of ignorance.

Irreducible  uncertainties can be “dangerous” when the developed  models are used for guiding decision making. They might lead to poor model prediction performance (which might be accurate but wrong), underestimation of the overall output uncertainty, and poor inferences on future health conditions,  ultimately resulting in  wrong decision making under uncertainty. 
To clarify further this point and the importance of accounting for uncertainty, in Figure 4, it is schematically shown a graph considering the influence of uncertainty (low, medium, and high) on a performance metric used to assess the behaviour of the dynamical system, versus the availability of information that can help reduce or properly quantify uncertainty.  

\begin{figure}[h]
\centering
\includegraphics[width=31pc]{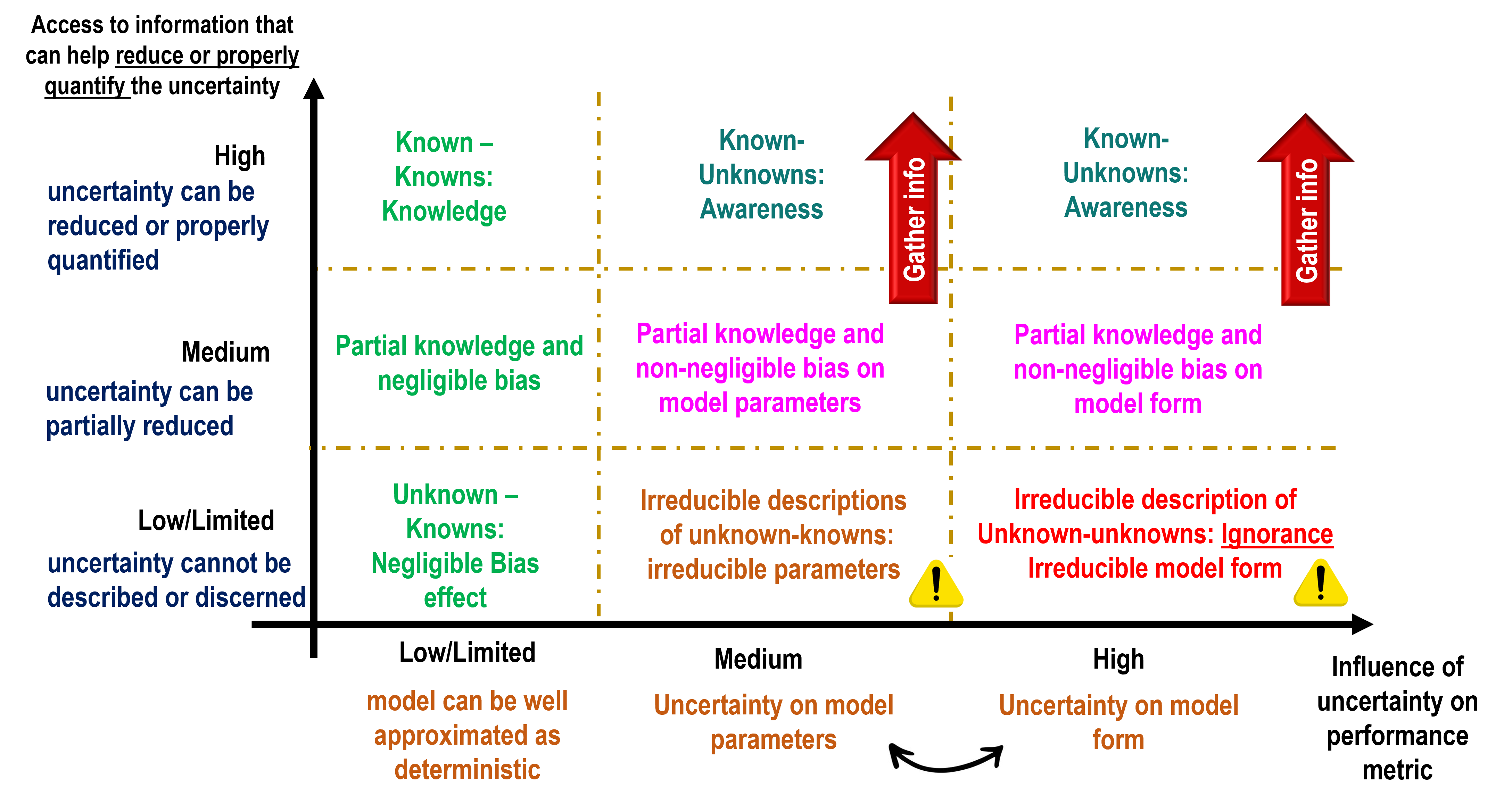}
\caption{Influence of uncertainty on a performance metric versus availability of information to reduce or quantify uncertainty. Depending on the application medium and high influence of uncertainty columns might be swapped.}
\end{figure}
When the influence of the uncertainty on the performance metric is low/limited, the model can be approximated as deterministic. If this influence is medium/high, and there is access to the information to reduce or properly quantify uncertainty, the uncertainty can be properly quantified to guide decision making. However, one might be aware of having partial availability of information and a non-negligible effect of the uncertain model parameters and/or model form. This would result in the decision to gather more information to properly reduce or quantify uncertainty. However, in the presence of irreducible model parameters or model form, the uncertainty cannot be properly quantified/reduced, significantly affecting the decision making. 

\subsection{Challenge four: Explainable and Interpretable inferences}
The concepts of interpretability and explainability introduced in the ML community (and presented comprehensively in \cite{refIntExp}) can be also extended to engineering models which are primarily used for guiding decision making of dynamical systems. 

\textit{Interpretability} focuses on being able to understand why the model would succeed/fail to complete a specific task, for example the prediction of an output variable.  The importance of a model being interpretable is related to the fact that humans would trust more predictions made by a so-called ``white-box" model (physics-based models), where physics assumptions, model approximations and causality are inherently visible, rather than the so-called ``black-box" models (data-driven model). Indeed, the ``failure" of a data-driven model can be related to causes that might be difficult to immediately identify: (i) sub-optimal model (e.g., local minima identification, underfitted/overfitted model, to name some); (ii) corrupted or non-informative measurements; (iii)  correlation and associational evidence that cannot be generalised; (iv) presence of confounding influences not being properly disentangled. Instead, the ``failure" of a white-box model can be potentially directly linked by the modeller to incorrect physical model parameters or incorrect modelling assumptions (assuming a LTI system). 

\textit{Explainability} focuses on communicating the predictions to support critical decision making. To be used in Engineering, predictions must be not only comprehensible (understandable by humans), but capable of quantifying any remaining uncertainty in the prediction to ensure the safe operation and the strategic planning of activities (e.g., inspections and maintenance).

\section{Physics-Enhanced Machine Learning (PEML)}
PEML provides a framework for guiding high-consequence decision making in engineering applications involving complex dynamical systems by developing hybrid physics-data models.  These hybrid models integrate advanced computational models (e.g., multi-fidelity, multi-scale, high-dimensional, coupled) and data (e.g., small, heterogeneous, gappy, noisy, multimodal - in the form of images, time series, lab test, historical data, inspection documents - and multi-fidelity, with different spatial and temporal resolution and quality), with (i) domain knowledge; (ii) prior knowledge (observational, empirical, physical, mathematical); and (iii) first principles and appropriate biases (e.g., physics-constraints, model form). PEML aims at: (i) overcoming poor generalization performance and physically inconsistent/implausible predictions; (ii)  accounting for and quantifying the different sources/types of uncertainties; (iii) providing explainable and interpretable inferences; to enable algorithms for:
\begin{itemize}
\item {\bf informing the physics describing the underlying dynamical system} to be able to analyze, control and predict the wide range of behaviours of the real-world system. This includes the identification of prediction models (e.g. estimating outputs at {\rm t+1} for control applications and virtual sensing), of forecasting models (e.g., estimating system behaviour in short-medium term for maintenance or SHM applications), of   simulation models (e.g., digital twin to investigate the behaviour of the system for new  inputs and/or to quantify uncertainty evolution in the model parameters), of the governing equations, of reduced order models, of multi-physics interactions, of unknown constitutive laws, of physics-based model parameters (also known as probabilistic model updating), of states.
\item 
{\bf fast and accurate solutions} of hybrid physics-data models, including governing equations, reduced order models, prediction,  forecasting and simulation models. 
\end{itemize}

\section{Physics and domain knowledge biases to obtain a PEML architecture}
 Biases allow a learning algorithm to prioritize one model and/or interpretation over another, even if not strongly supported by the data. The role of biases is to restrict the region of the admissible solutions that will be explored. The need of biases to yield an``inductive leap" and make a ML algorithm learn ``one generalization over another, other than strict consistency with the observed training instances''' was proposed in \cite{Mitchell}. In principle, different types of biases can be combined to accelerate training and enhance generalisation even with limited data.  Typically, in ML, all biases are labelled as ``inductive", and alternative classifications of biases have been proposed in the context of PEML \cite{ref2, eleni, ref11}.  Based on the work in \cite{ref11}, it is proposed to consider four types of physics and domain knowledge biases, schematically summarised in Figure 5:
\begin{itemize}
    \item  \textbf{Observational bias}: Introducing data that embody underlying physics, i.e., informative data. The data can be synthetically generated, and used to augment measurements.
    \item  \textbf{Learning bias}: Selecting a learning model that helps us getting faster to the underlying model description by enforcing biases within the inference/learning algorithm itself. This includes, but is not limited to: selection of the algorithm itself, loss functions, optimization setup, learning algorithm constraints, numerical integration schemes.
    \item  \textbf{Inductive bias}:   Modifying the algorithm architecture,  by incorporating prior assumptions and physical constraints that we know of. This include, but is not limited to: conservation laws, symmetries, invariances, boundary conditions, library of candidate functions. 
    \item  \textbf{Model form/discrepancy bias}. This corresponds to including “terms” from partial knowledge of the physics-based model (with or without uncertainty) - for example part of the known governing equation - and/or assuming a specific physics-based model form. 
\end{itemize}
\noindent The last bias is particularly important for dynamical systems since a partial knowledge of the physics-based model is usually available (e.g., \cite{ref4,ref4bis,eleni, Zou}).  The proposed distinction of four biases is not limited to a specific type of hybrid physics-data model (e.g., PINNs \cite{ref2} or Physics-Enhanced Sparse System Identification of Dynamical Systems (PhI-SINDy) \cite{ref11}), and it is independent on how the biases are going to be implemented. For example, within PINNs, the inductive and model form/discrepancy biases can be  implemented into the loss function \cite{ref2}. In PhI-SINDy \cite{ref11}, inductive biases can be introduced in the library of candidate functions, and by setting physics constraint in the numerical integration scheme. 
\begin{figure}[htb!]
\centering
\includegraphics[width=30pc]{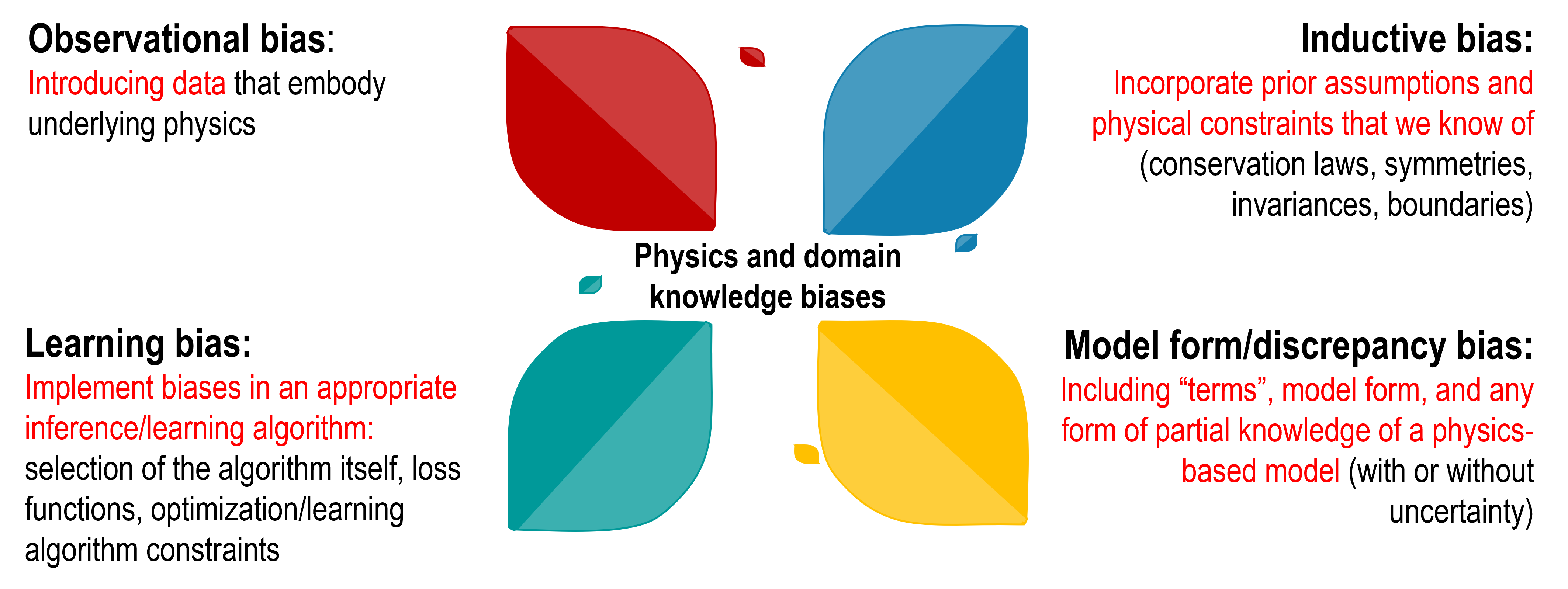}
\caption{Summary of the four categories of biases.}
\end{figure}
\vspace*{-\baselineskip}
\begin{figure}[htb!]
\centering
\includegraphics[width=20pc]{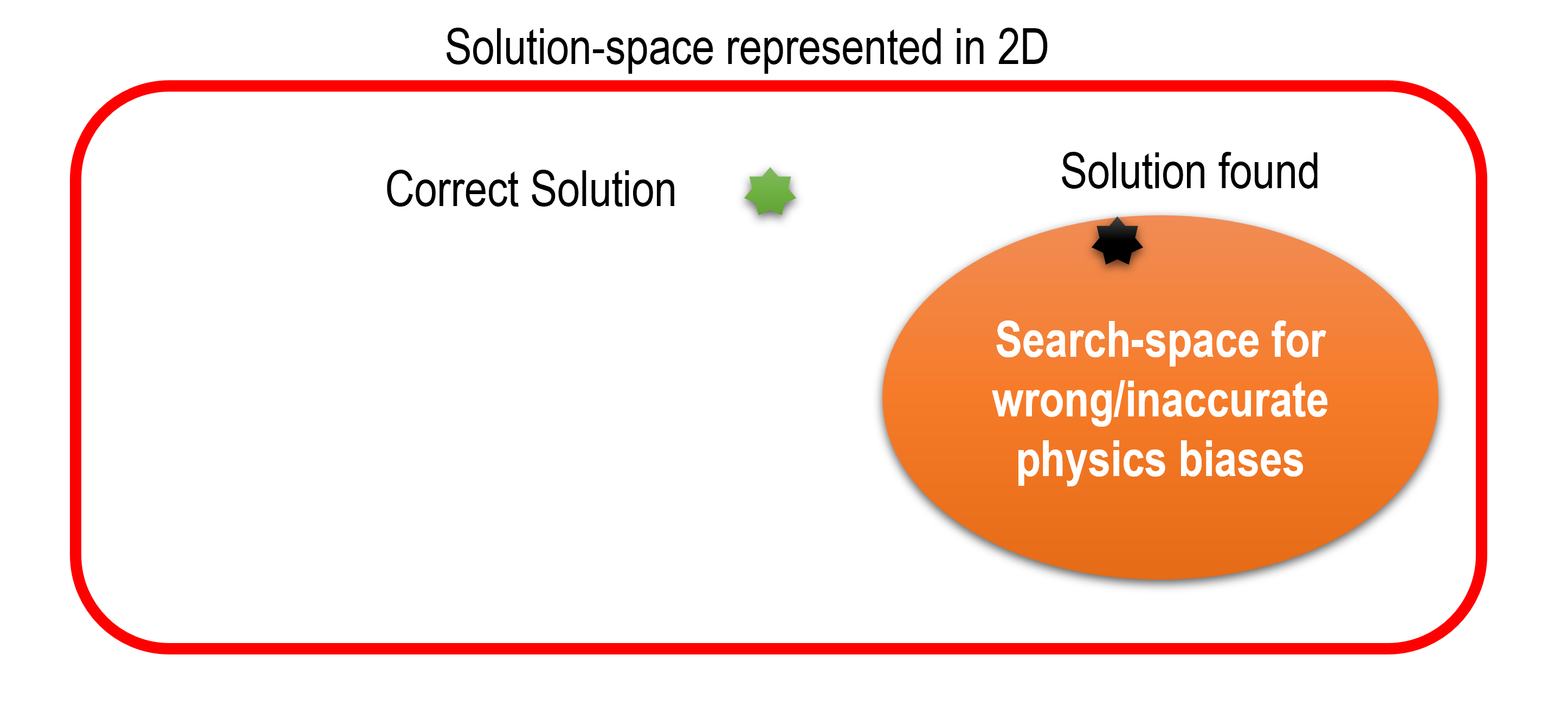}
\caption{Wrong/inaccurate physics bias effect: reducing the solution-space to a wrong region.}
\end{figure}
\vspace*{-\baselineskip}

 Since the role of these biases is to restrict the region of the physics-space that will be explored, if a perfect knowledge of the underlying physics of a deterministic model was available, the biases would result in the identification of a point, i.e., the ideal model. The presence of uncertainty would lead to the identification of a region.  This interpretation helps highlighting a key challenge in a PEML model: introducing one or more inaccurate or wrong biases (i.e., not representative of the problem under investigation) can severely undermine the accuracy of the PEML strategy implemented, since entire valid regions of the solution space would remain unexplored, and a sub-optimal or wrong model would be identified. 
This it schematically summarised in Figure 6.

\subsection{Three broad groups of PEML approaches}
 PEML strategies can be grouped as summarised in Figure 7. These groups are based on  building a PEML model ($M$) by answering two main questions: (i) how much physics vs data should be included?;  (ii) how are  physics, data and physics and domain knowledge biases integrated?

\begin{figure}[h]
\centering
\includegraphics[width=30pc]{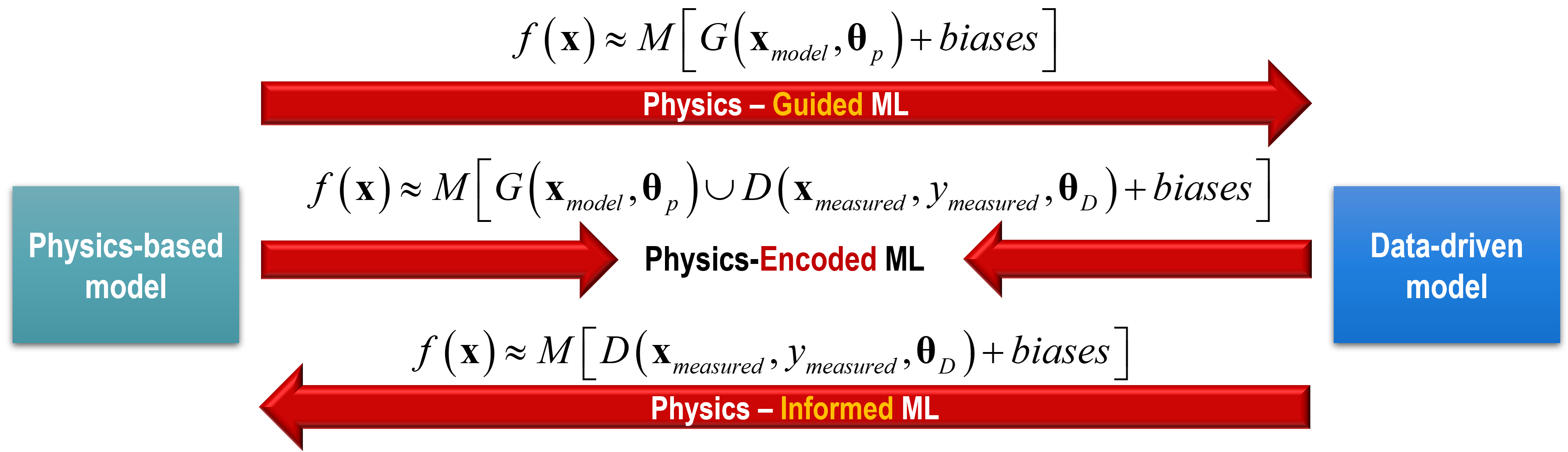}
\caption{Physics-Guided ML, Physics-Encoded ML and Physics-Informed ML.}
\end{figure}

\subsubsection{Physics-Informed ML: $f({\bf x})\approx M[D({\bf x}_{measured}, {y}_{measured}, \boldsymbol{ \theta}_D)+biases)]$}
Medium/large volume of informative data is available. A data-driven model $D$ with parameters $\boldsymbol{ \theta}_D$ is developed, and biases are introduced to constrain the space of admissible solutions that are going to be explored.   For example, measured data can be replaced by synthetic data obtained with high-fidelity physics-based models (observational bias) to develop reduced order models. Biases can be introduced in the loss function in PINNs architectures to accelerate a Partial Differential Equation solver \cite{ref2}. Biases can be introduced by incorporating a numerical integration scheme (learning bias) \cite{ref1}, or stick-slip temporal constraints \cite{saurabh} in sparse system equation discovery \cite{ref3}. Biases can be introduced by selecting physics-based mean and kernels, and constraints, of Gaussian Process (GP) regression models \cite{ref4, ref4bis}. It is worth noting that given the proposed definition, Physics-Guided NNs, Physics-Informed NNs and Physics-Encoded NNs (e.g., \cite{ref5}) can be all classified as Physics-Informed ML strategies. 

\subsubsection{Physics-Guided ML: $f({\bf x})\approx M[G({\bf x}_{model}, \boldsymbol{ \theta}_p) + biases]$ }
A detailed physics-based model $G$ is believed to be highly accurate and representative of the dynamical system being investigated. Small/medium informative measurements (data) are available (observational bias). Additional biases can be also introduced.  Typically, the focus of these approaches is the identification of the latent physics-based model parameters $ \boldsymbol{ \theta}_p$, of model prediction errors, input or joint input-state estimation. Examples include probabilistic model updating strategies (see for example \cite{eleni, KH, lye}). It is worth mentioning that introducing biases for improving the performance of a detailed physics-based model, especially in forecasting tasks,  significantly  increase the computational cost.   As an example, let us consider an application of Physics-Guided ML to real-world dynamical systems \cite{ref10}: a twin girder bridge for which a detailed physics-based model is available and multiple measurements along the span are provided to identify several latent parameters of the model. An approach for the Bayesian system identification of structures considering spatial and temporal correlation was developed explicitly accounting for competing spatial and temporal correlations models (inductive bias). To be feasible, an efficient loglikelihood calculation procedure was developed \cite{ref10}. 

\subsubsection{Physics-Encoded ML: $f({\bf x})\approx M[G({\bf x}_{model}, \boldsymbol{ \theta}_p) \cup D({\bf x}_{measured}, {y}_{measured}, \boldsymbol{ \theta}_D)+ biases]$}

These approaches build models $M$ which explicitly combine data-driven $D$ and physics-based $G$ components. This type of models are typically used when it is believed that the physics-based model (which might have uncertain parameters) is not capable of representing fully the underlying physics of the problem and small/medium informative data is available to learn the missing parts of the physics (and the uncertain parameters of the known physics-based part). Biases are then introduced to constrain the space of admissible solutions that are going to be explored. Examples include GP Latent Force Models (e.g., \cite{Zou,Marino}), PhI-SINDy \cite{ref11}, and others can be found in \cite{ref4,ref4bis,eleni}.  For example, for the identification of a frictional contact in laboratory conditions \cite{Marino}, the unknown frictional force of a known Single Degree of Freedom model (with uncertain parameters) can be modelled as a GP to be learned through a Switching GP Latent Force model enhanced by introducing inductive, learning and model form biases.

\section{Which PEML strategy should be used?}

There is no one-size-fits-all PEML strategy. The choice is highly dependent on three aspects:
\begin{itemize}
    \item \textit{Available information}: The amount of physics and domain knowledge on the dynamical system vs informative data that should be included in a PEML model, depends on the specific application. The model developer should be carefully assessing the quality of the available information when deciding to adopt physics-guided, physics-informed or  physics-encoded approaches. This selection should not be based on the modeller preference or previous experience on a different problem. 
    \item \textit{Users purpose}: Depending on the task at hand, different PEML strategies might be more appropriate. For example, for accelerating a solver, Physics-Informed ML strategies should be preferred (see for example \cite{ref2, ref4,ref4bis,eleni}). If the objective is the identification of unknown physics/terms of a model, approaches based on sparse regression strategies (see for example \cite{ref3, ref1, ref11}) might be more effective. When dealing with the development of Digital Twins \cite{DTpaper} of a time-evolving digital model that can account for uncertainty,  different systems interactions, and that can be used for policy and decision making, physics-guided or physics-encoded approaches might be more appropriate.  
    \item \textit{Complexity of the system/problem}: Approaches very effective for handling real-world systems that can be approximated as LTI systems (e.g., \cite{Zou, eleni}), might not be well suited for dealing with nonlinear time-varying system (e.g., case of discontinuous nonlinearity caused by frictional contacts  \cite{Marino}). Tackling the complexity of Multi-scale, Multi-physics, Nonlinear time-varying dynamical systems is still beyond state-of-the-art PEML strategies. 
\end{itemize}
\vspace*{-\baselineskip}

\section{Open challenges and opportunities with particular focus on dynamical systems}

\begin{itemize}
    \item Metrics  to assess prediction quality, benchmarks and validation strategies. Different PEML algorithms might return the same performance on training data, however their performance on unseen data (test data) might be very different, as well as their ability to be generalizable. There is a need to develop metrics, complex benchmark case studies and validation strategies that can be used to rate the performance of different PEML algorithms on specific tasks. 
    \item  Automatic identification and correction of errors in the data, physics and domain knowledge biases and/or in the chosen PEML model architecture. Including detecting errors caused by: (i) wrong modelling assumptions and/or biases (e.g., wrong constitutive model); (ii) corrupted data; (iii) non-informative data; (iv) wrong architecture setup where the data-driven part incorrectly overrides the physics-based part \cite{overrading_phy}; (v) learning a discrepancy which is not generalizable to unseen conditions; (vi) wrong uncertainty quantification. 
  \item Scalable solutions for complex problems. Including: (i) mapping in a low-dimensional space to speed-up computation while retaining accuracy; (ii) high-precision learning from small informative datasets; (iii) dealing with the presence of non-smooth nonlinear phenomena; (iv) efficiently quantifying uncertainties and their impact on decision making.
\end{itemize}
Most importantly, developing better understanding of complex problems (e.g., how to model nonlinearity and time-varying phenomena) and correctly quantifying uncertainty are necessary steps to move from accurate-but-wrong predictions, to explainable and interpretable inferences.









 

\vspace*{-\baselineskip}

\end{document}